\documentclass[
]{ceurart}

\usepackage{listings}
\usepackage{graphicx}
\usepackage{comment}
\usepackage{todonotes}
\usepackage{tikz}
\usepackage{caption}
\usepackage{subcaption}
\usepackage{varwidth}
\usepackage{url}
\begin{document}

\copyrightyear{2022}
\copyrightclause{Copyright for this paper by its authors.
  Use permitted under Creative Commons License Attribution 4.0
  International (CC BY 4.0).}

\conference{AIC 2022, 8th International Workshop on Artificial Intelligence and Cognition}

\title{Towards Learning Abstractions via Reinforcement Learning}

\author[1]{Erik Jergéus}[%
orcid=0000-0003-2231-6869,
email=erikjer.student@chalmers.se
]
\cormark[1]
\fnmark[1]

\author[1]{Leo Karlsson Oinonen}[%
orcid=0000-0003-4117-5096,
email=leoo.student@chalmers.se,
]
\cormark[1]
\fnmark[1]

\author[1]{Emil Carlsson}[%
orcid=0000-0002-0170-7898,
email=caremil@chalmers.se]

\author[1]{Moa Johansson}[%
orcid=0000-0002-1097-8278,
email=jomoa@chalmers.se]

\address[1]{Chalmers University of Technology, Gothenburg, Sweden}

\cortext[1]{Corresponding author.}
\fntext[1]{These authors contributed equally.}

% % \author{Erik Jergéus\inst{1}\orcidID{0000-0003-2231-6869} \and
% % Leo Karlsson Oinonen\inst{1}\orcidID{0000-0003-4117-5096} \and
% % Emil Carlsson\inst{1}\orcidID{0000-0002-0170-7898} \and
% % Moa Johansson\inst{1} \orcidID{0000-0002-1097-8278}}
% %
% %\authorrunning{F. Author et al.}
% \authorrunning{E. Jergéus, L. Karlsson Oinonen, E. Carlsson, M. Johansson}
% \institute{Chalmers University of Technology, Gothenburg, Sweden\\ 
% %\url{https://www.chalmers.se/} \and
% \email{\{erikjer.student, leoo.student, caremil, jomoa\}@chalmers.se}}
% %ABC Institute, Rupert-Karls-University Heidelberg, Heidelberg, Germany\\
% %\email{\{abc,lncs\}@uni-heidelberg.de}}
% %
%
\begin{abstract}
In this paper we take the first steps in studying a new approach to synthesis of efficient communication schemes in multi-agent systems, trained via reinforcement learning. We combine symbolic methods with machine learning, in what is referred to as a neuro-symbolic system.
The agents are not restricted to only use initial primitives: reinforcement learning is interleaved with steps to extend the current language with novel higher-level concepts, allowing generalisation and more informative communication via shorter messages. We demonstrate that this approach allow agents to converge more quickly on a small collaborative construction task.

% This paper investigates construction of a Domain Specific Language for use in multi-agent neural networks. The language aims to be both efficient in the amount of communication needed and result in high interpretability. Much like natural language, the language should achieve this through abstracting common ideas. Our method results in X percent faster learning for biased samples, but impedes learning when used in randomized environments. However, in every case a trained network requires less information transferral, when utilizing abstractions.

%The abstract should briefly summarize the contents of the paper in 15--250 words.

\end{abstract}


\maketitle

\section{Introduction}

Learning to communicate and coordinate efficiently via interactions, rather than relying on solely supervised learning, is often viewed as a prerequisite for developing artificial agents able to do complex machine-to-machine and machine-to-human communication \cite{Mikolov18}. The field of language learning and emergent communication has a long history \cite{Hashimoto1996EmergenceON,kirby&hurford2002,Smith03,steels2005emergence,Steels2015}, and is now a vibrant field of research also in the deep learning community \cite{Foerster16,lazaridou2020emergent,Hill&etal2020,Hill&etal2020a}. Recent work has focused on developing agents with single message communication \cite{Jorge2016,Lazaridou2017,Kageback2020}, variable length communication \cite{Havrylov17} and compositional language \cite{Mordatch18, mu2021emergent}, via interactions and reinforcement learning. However, a striking characteristic of human communication that has been overlooked in the literature is the ability to derive novel concepts and abstractions from primitives, via interaction. 

\begin{figure}[h]
    \centering
    \includegraphics[scale=0.4]{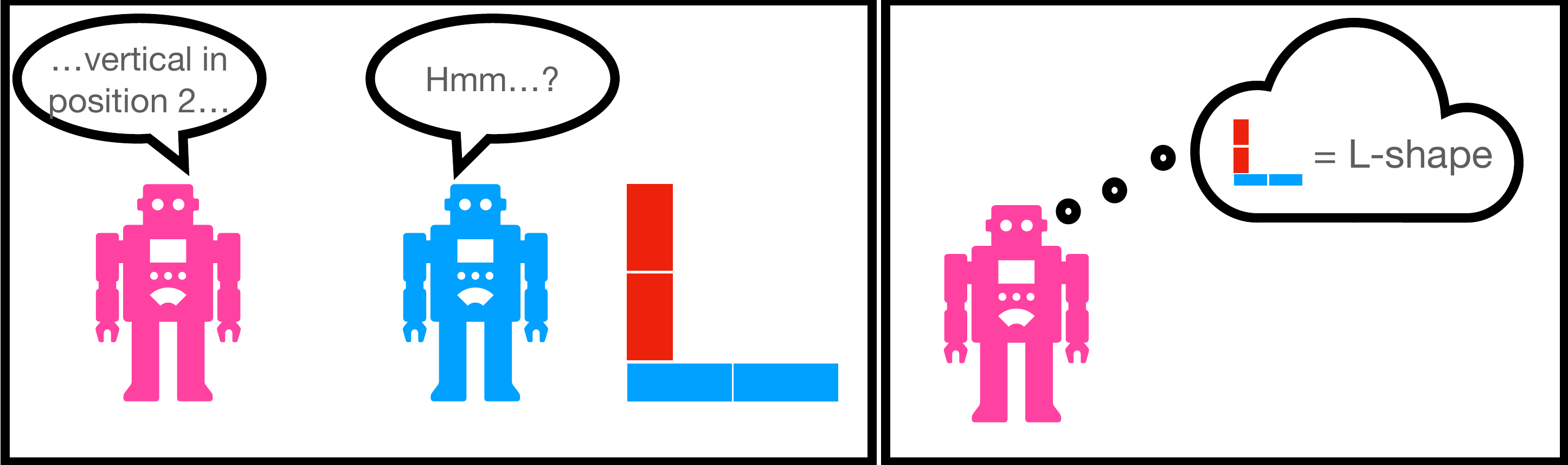}
    \caption{Agents should periodically reflect on their experience and consider introducing abstractions, allowing shorter utterances for constructing commonly occurring shapes.}
    \label{fig:archbuild}
\end{figure}

In this paper, we investigate how artificial agents can develop linguistic abstractions via interaction and reinforcement learning, starting from a small set of primitive concepts and gradually increasing the size and efficiency of their language over time.
Our motivation is the builder-architect experiment in \cite{McCarthyBuilding21}, investigating how humans develop communicative abstractions. Here, the architect is given a drawing of a shape, and has to instruct the builder how to construct it from small blocks. %humans develop abstractions via interaction when given the task to construct shapes out of small blocks. 
As the experiment progressed, participants developed more concise instructions after repeated attempts. Instead of talking about the positions of individual blocks, they started using abstractions describing commonly seen shapes, such as \emph{L-shape} or \emph{upside-down U}, see Figure \ref{fig:archbuild}.  
Our contribution here is an initial feasibility study of a neuro-symbolic multi-agent reinforcement learning framework for this task. Inspired by neuro-symbolic program synthesis \cite{DreamCoder2021}, the agent interleave reinforcement learning to train their neural network, with symbolic reflection to introduce new concepts for common action sequences. We show that agents learn to reconstruct the given shapes faster when allowed the capability to introduce abstractions.

\section{Implementation}
Our setup mimics the one from McCarty et al. \cite{McCarthyBuilding21} where two agents, the architect and the builder, communicate about a set of geometric shapes. %Agents are modelled as neural networks. %In our case, we model each agent as neural networks. 
The agents iterate between two learning phases, one where the agents use reinforcement learning to develop the meaning of each message, followed by an abstraction phase, where the architect may introduce new instructions for commonly seen structures. 
This will allow the agents to potentially solve the tasks using fewer messages, which gives them a higher reward as shorter interactions are preferred.

%gain even higher rewards in the next round, as fewer messages are used to solve the task, and our setup gives higher rewards to shorter interactions. %they can solve the task using fewer interactions.

\subsection{The Environment}
\begin{figure}[htbp]
    \centering
    \includegraphics[width=0.7\textwidth]{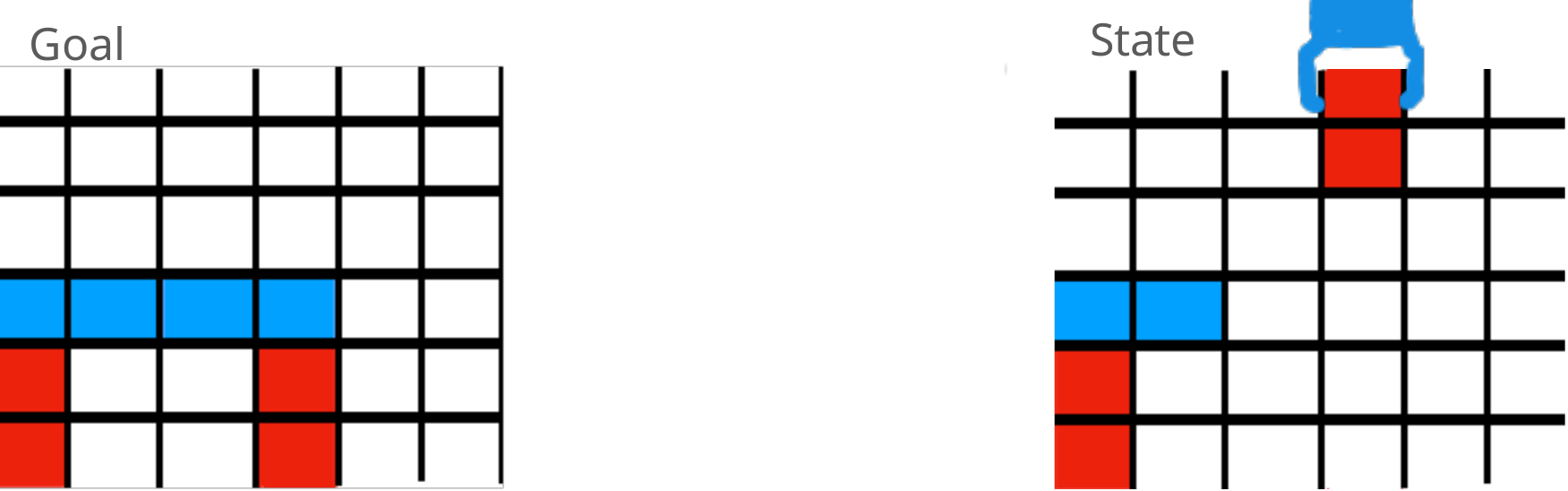}
    \caption{The architect sees both the goal and the current state and decides to instruct the builder to place a vertical block in position 4.}
    \label{fig:place_block}
\end{figure}

The architect's input is a picture of the goal state alongside the current state, each of which is represented by binary 6x6 matrices, see Figure \ref{fig:place_block}. Locations where there are blocks are represented as 1's and empty locations by 0. This is passed through a feed forward neural network, which outputs a message with instructions to the builder. 
The work described here focus on the learning of the architect, with the builder assumed to understand the architect's messages perfectly. In the future, the builder will also be represented with a neural-network, learning to map messages to the corresponding (sequence of) actions.
Initially, the architect's message-space consists of 12 messages, simply the six possible positions of vertical (2 x 1) and horizontal (1 x 2) blocks respectively. We denote the set of messages as $\mathcal{M}=\{V_1, H_1, ..., V_6, H_6\}$.
These initial messages have a one-to-one correspondence to the basic actions the builder can perform. Note that the architect is allowed to introduce new messages, abstractions, during the abstraction phase, formally introduced in later sections. 
\paragraph{Reward Function}
The agents receive a reward $R$ at each time step $t$ when performing an action $a$, either a larger reward if the new state matches the goal exactly, or a smaller reward if the most recently placed block partially matches the goal.
This reward function is given in Equation \ref{eq:architect_reward}, where \emph{partial\_match} denote the number of new grid squares covered by the most recently placed block matching the goal. 
\begin{equation}
R_t(s,g,a) = (0.1 * partial\_match + 1 * (s == g)) * 0.9^{t}
\label{eq:architect_reward}
\end{equation}
The architect becomes better at generalising what placing a single block entails when receiving an intermediate reward based on how much of that block contributes to the final goal. The larger reward from completing the whole structure biases the architect to always aim for a perfect completion of the goal. 
Finally, we encourage the architect to always use as few messages as possible (i.e. using abstractions) by discounting the reward based on the number of time-steps further.

\subsection{Deep Reinforcement Learning}
The architect is modelled as a Deep Q-Network (DQN) with experience replay \cite{Mnih13}. In short, this means that the architect, for each state and message, $(s, m)$, estimates the Q-value, or expected cumulative reward, for conveying message $m$ given state $s$. We consider a neural network with layers of sizes
$[72, 576, 576, 576, 36, |\mathcal{M}_{max}|]$, where $\mathcal{M}_{max}$ is the maximum allowed size of the message space, and we use ReLU activation between each layer. See GitHub\footnote{\url{https://github.com/jerge/MARL/tree/Communicative-Abstractions}} for the other hyper-parameters and implementation.

\subsection{Abstraction Phase}
 In order to learn abstractions, we implement a version of the wake-sleep-dream framework used in the neuro-symbolic system DreamCoder \cite{DreamCoder2021}. % which consists of a sleep and dream phase. 
 After a \emph{wake-phase} where the agents have engaged in reinforcement learning to learn to communicate using the current set of messages, the architect enters the \emph{sleep-phase}, where it is allowed to invent new messages. During the sleep phase, the architect searches for the longest common sub-sequence(s) of messages from the previous reinforcement learning phase. The sub-sequence's are rated based on their length and frequency and the top-rated sequence will turn into a new abstraction. 

Next follows a \emph{dream phase}, which aims to quickly train the agents to use the new abstraction generated from the sleep phase. This is done by letting the agent re-experience the examples from the replay buffer, but now with the new abstraction instead of the corresponding sub-sequence of messages. This will lead the new abstraction towards the appropriate Q-value before starting the next reinforcement learning phase.

\section{Experimental Results}
\begin{figure}[htpb]
    \centering
    \includegraphics[scale=0.3]{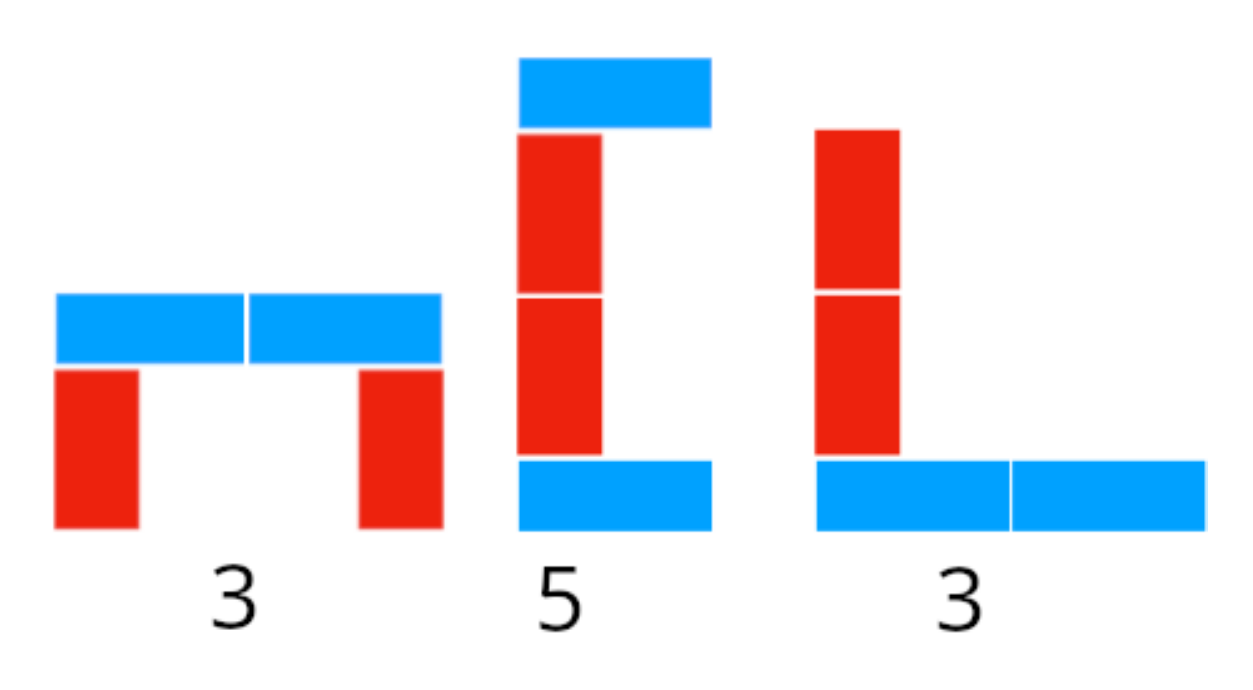}
    \caption{Shapes in our experimental set: 3 \emph{upside-down U}, 5 \emph{C}-shapes and 3 \emph{L}-shapes in all possible different locations in the 6x6 grid.}
    \label{fig:structured}
\end{figure}

%In order to assess if our framework is a feasible setup for the abstraction task, we ran an experiment:
We conducted an initial feasibility experiment for our framework: Given a set of three re-occurring shapes from McCarthy's human experiment (Fig. \ref{fig:structured}) \cite{McCarthyBuilding21}, does the agents learn to reconstruct them faster if allowed to introduce abstractions? 

We hypothesise that:
% a) having a language with messages also corresponding to common sequences of actions will facilitate the reinforcement learning construction task, and b) our neuro-symbolic agent can discover and learn to use such concepts.
\begin{description}
\item[a)] Having a language with messages also corresponding to common sequences of actions will facilitate the reinforcement learning construction task.
\item[b)] Our neuro-symbolic agent can discover and learn to use such concepts.
\end{description}

\begin{figure}[htpb]
     \centering
     %with and w/o builder
     %\begin{subfigure}[t]{0.49\textwidth}
         \centering
         \includegraphics[scale=0.52]{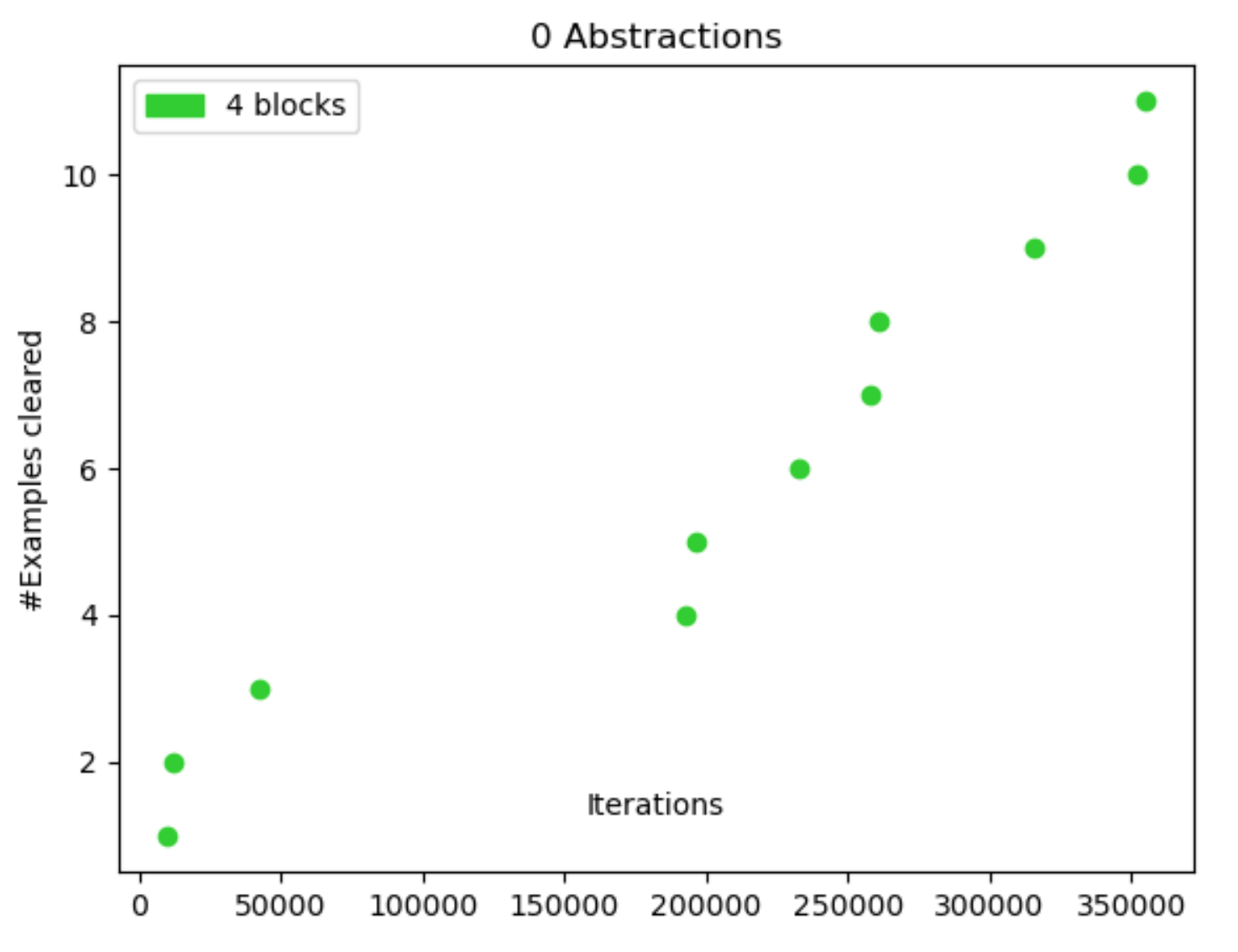}
         \caption{Worst-case bound: without the capability to create abstractions, using only initial primitives, learning to build all shapes requires over 350 000 epochs.}
         %This graph shows the median number of iterations needed for the agents to clear each construct in the set $S_{structured}$, when the networks were capped at 0 abstractions.}
         \label{fig:worst-case}

\end{figure}
\begin{figure}[htbp]%{0.49\textwidth}
         \centering
         \includegraphics[scale=0.52]{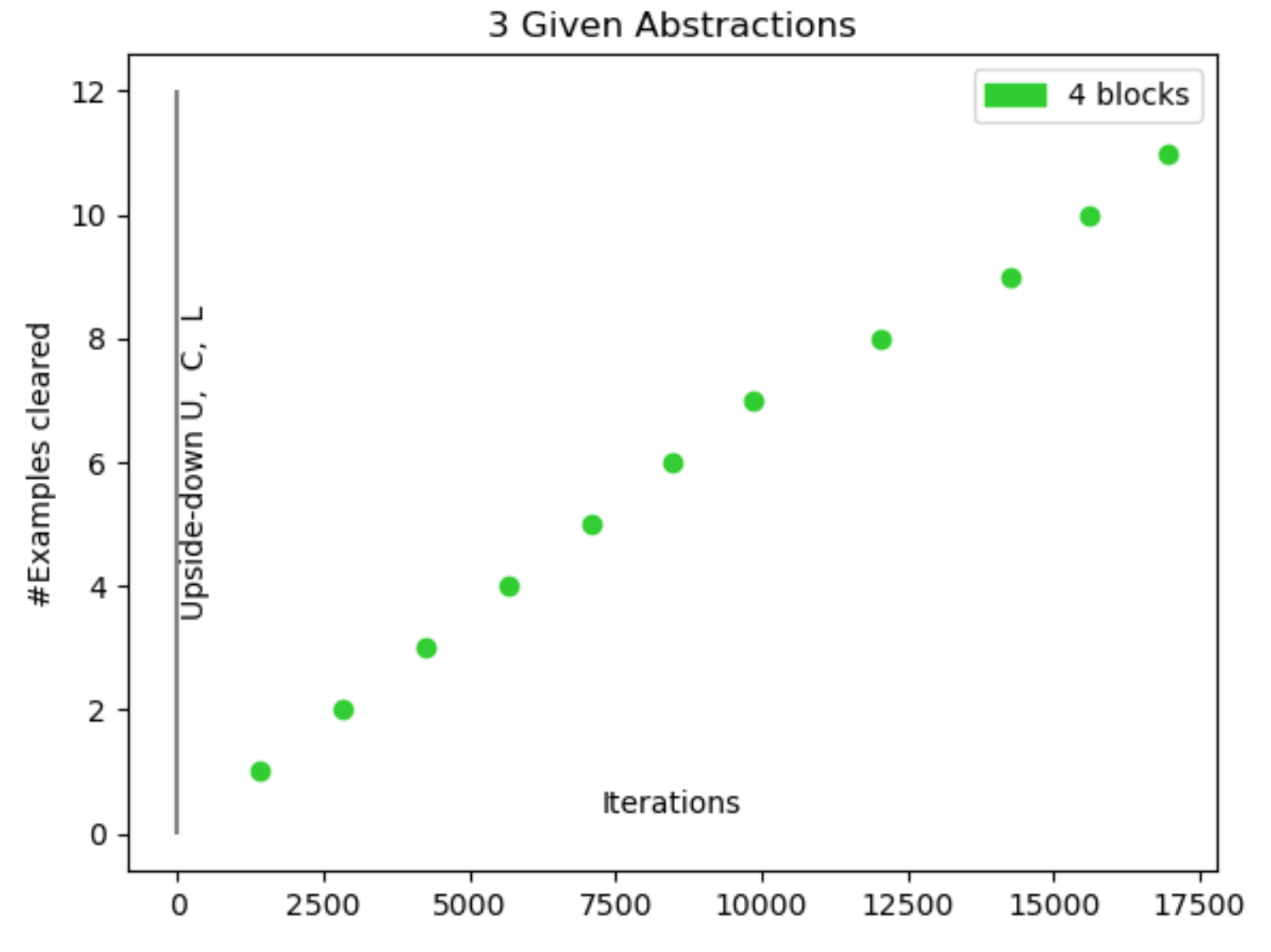}
         \caption{Best-case bound: If agents are given the relevant abstractions upfront, the task can be learned in 17 500 epochs.
         %This graph shows the median number of iterations needed for the agents to clear each construct in the set $S_{structured}$, when the networks were given the abstractions corresponding to each of the structures in $S_{structured}$.
         }
         \label{fig:best-case}
\end{figure}
\begin{figure}[htbp]
\centering
         \includegraphics[scale = 0.52]{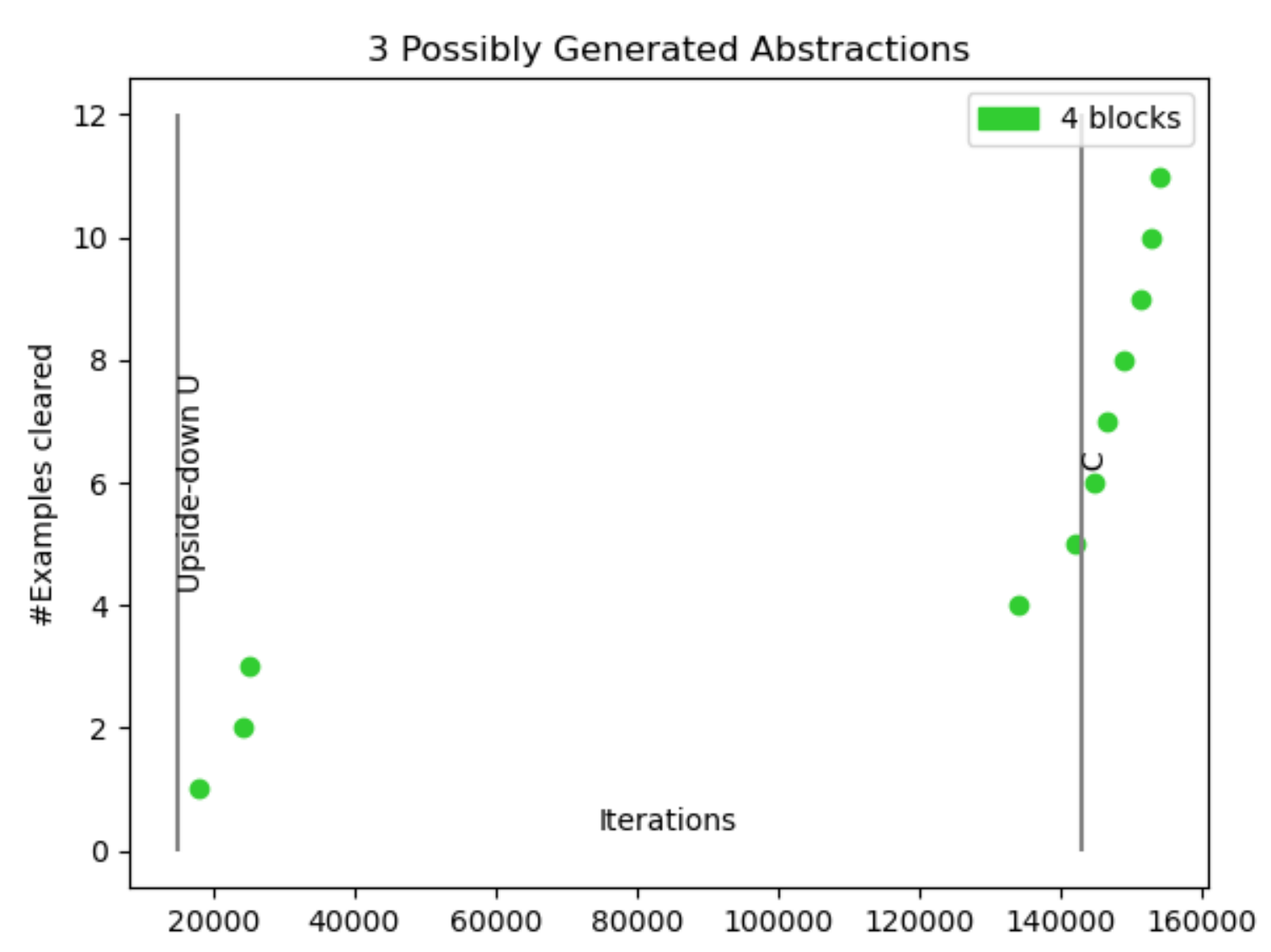}
         \caption{With abstraction-invention, the agents need 160 000 epochs to learn to build all shapes. The horizontal lines mark when the abstraction \emph{upside-down U} and \emph{C}-shape were introduced. }
         %This graph shows the median number of iterations needed for the agents to clear each construct in the set $S_{structured}$, when the networks were capped at 3 abstractions. An 'Upside-down U' was generated around iteration #15032 and a C was determined around epoch #143050 and after that the algorithm did not need to make any abstractions.}
         \label{fig:with_abstraction}
\end{figure}  

Agents were first pre-trained on simple randomly generated shapes to learn the basic message/action pairs.
To establish an upper and lower bound on the agents' performance, we evaluated one instance without abstraction capabilities (worst case, Fig. \ref{fig:worst-case}), and one instance where optimal abstractions for the shapes were already given upfront (best case, Fig. \ref{fig:best-case}). The differences are large: in the worst case the agents required 350 000 epochs to successfully learn to construct all shapes, compared to 17 500 epochs in the best case. There is a clear advantage in having a richer language.

% \begin{figure}[h]
%     \centering
%     \begin{tabular}{cc}
%     \multicolumn{2}{c}{
%         \subfloat[Worst-case bound: without the capability to create abstractions, using only initial primitives, learning to build all shapes requires over 350 000 epochs.\label{fig:worst-case}
%          ]{\includegraphics[scale=0.4]{biased_0_success.png}} 
%     }
    
%     \\
%     \subfloat[Best-case bound: If agents are given the relevant abstractions upfront, the task can be learned in 17 500 epochs.\label{fig:best-case}
%          ]{\includegraphics[scale=0.4]{biased_3_prior(2).png}}    &
%     \subfloat[With abstraction-invention, the agents need 160 000 epochs to learn to build all shapes. The horizontal lines mark when the abstraction \emph{upside-down U} and \emph{C}-shape were introduced.\label{fig:with_abstraction}]{\includegraphics[scale = 0.4]{biased_3_gen(2).png}}         

%     \end{tabular}
%     \caption{Experimental results}
% \end{figure}

Next, we evaluated the complete architect-agent with ability to introduce abstractions, as shown in Fig. \ref{fig:with_abstraction}. The agents learned to solve the construction task after 160 000 epochs, which is still considerably faster than the worst-case scenario. Note that the architect choose to introduce only two abstractions, the \emph{upside-down U} very early on, and the \emph{C}-shape towards the end. Investigating this behaviour in more detail is further work.
    
\section{Conclusion and Future Work}
In this work, we have introduced a neuro-symbolic framework for learning linguistic abstractions via a combination of reinforcement learning, symbolic reasoning and interactions between agents. Our initial results on a small collaborative building task suggest that it is feasible for reinforcement learning agents to develop useful abstractions by alternating between neural learning, and symbolic abstraction phases introducing new concepts. These introduced abstract concepts also greatly improve the performance of the agents.

This is just a first step and we would like to further explore how to learn abstractions via reinforcement learning. One interesting direction is to extend our work to more complex environments. One issue that might arise in such scenarios is that the agents might need to first develop several intermediate abstractions, before being able to construct abstractions that greatly improves the reward. Solving this type of exploration-exploitation dilemma seems like a fundamental problem for the agents, and might require new exploration techniques. 

Another interesting future direction is to explore scenarios where agents do not share exactly the same understanding of a message, and are required to reason about each other in a recursive fashion.

\paragraph{Acknowledgement}
Erik and Leo received a Lars Pareto Travel Grant from Chalmers University of Technology to present this work. Emil Carlsson was supported by CHAIR (Chalmers AI Research), and Moa Johansson was supported by the Wallenberg Al, Autonomous Systems and Software Program - Humanities and Society (WASP-HS) funded by the Marianne and Marcus Wallenberg Foundation and the Marcus and Amalia Wallenberg Foundation.

\bibliography{refs.bib}
% Other
% TEMPLATE

\end{document}